\documentclass{article}
\usepackage{spconf,amsmath,graphicx,hyperref}    

\usepackage{cite}
\usepackage{microtype}
\usepackage{amsmath,amssymb,amsfonts}
\newtheorem{lemma}{Lemma}

\usepackage{algorithmic}
\usepackage{graphicx}
\usepackage{textcomp}
\usepackage[dvipsnames]{xcolor}
\usepackage[ruled,vlined]{algorithm2e}

\title{BiRQ: Bi-Level Self-Labeling Random Quantization \\for Self-Supervised Speech Recognition}
\name{Liuyuan Jiang$^\star$, Xiaodong Cui$^\dagger$, Brian Kingsbury$^\dagger$, Tianyi Chen$^\ddagger$, Lisha Chen$^\star$
\thanks{This work was conducted while the first author was a visiting student at IBM Research through the RPI-IBM Future of Computing Research Collaboration. The work was supported by the National Science Foundation Projects 2401297 and 2412486, and by IBM through the IBM-Rensselaer Future of Computing Research Collaboration.}}
\address{$^\star$University of Rochester~~~~~~~~ $^\dagger$IBM Research~~~~~~~~   $^\ddagger$Cornell University}
\begin{document}
\ninept
\maketitle
%



\begin{abstract}
Speech is a rich signal, and labeled audio–text pairs are costly to obtain, making self-supervised learning (SSL) essential for scalable representation learning. A core challenge in speech SSL is generating pseudo-labels that are both informative and efficient: strong labels, such as those used in HuBERT~\cite{hsu2021hubert}, improve downstream performance but rely on external encoders and multi-stage pipelines, while efficient methods like BEST-RQ~\cite{chiu2022self} achieve simplicity at the cost of weaker labels.
We propose BiRQ, a bilevel SSL framework that combines the efficiency of BEST-RQ with the refinement benefits of HuBERT-style label enhancement. The key idea is to reuse part of the model itself as a pseudo-label generator: intermediate representations are discretized by a random-projection quantizer to produce enhanced labels, while anchoring labels derived directly from the raw input stabilize training and prevent collapse. Training is formulated as an efficient first-order bilevel optimization problem, solved end-to-end with differentiable Gumbel-softmax selection.
This design eliminates the need for external label encoders, reduces memory cost, and enables iterative label refinement in an end-to-end fashion. BiRQ consistently improves over BEST-RQ while maintaining low complexity and computational efficiency. 
We validate our method on various datasets, including 960-hour LibriSpeech, 150-hour AMI meetings and 5,000-hour YODAS, demonstrating consistent gains over BEST-RQ.


\end{abstract}
\begin{keywords}
self-supervised learning, bilevel optimization, BEST-RQ, random quantizer
\end{keywords}
\section{Introduction}

Large language models (LLMs) have shown remarkable effectiveness, enabled by abundant discrete text data~\cite{devlin2019bert}. In contrast, speech is continuous, and labeled audio–text pairs are scarce, making supervised auto-speech recognition (ASR) training challenging. Self-supervised learning (SSL) addresses this by discretizing signals into pseudo-labels (PLs) and training models to capture hierarchical information from low-level acoustics to higher-level lexical and semantic knowledge~\cite{baevski2020wav2vec,hsu2021hubert,chiu2022self}. These pretrained models can then be fine-tuned on downstream tasks like ASR with limited data, and their discrete representations can be aligned with text tokens via lightweight modules such as Q-formers for integration with LLMs~\cite{li2023blip}.

\begin{figure}[t]
\centering
\includegraphics[width=0.7\linewidth]{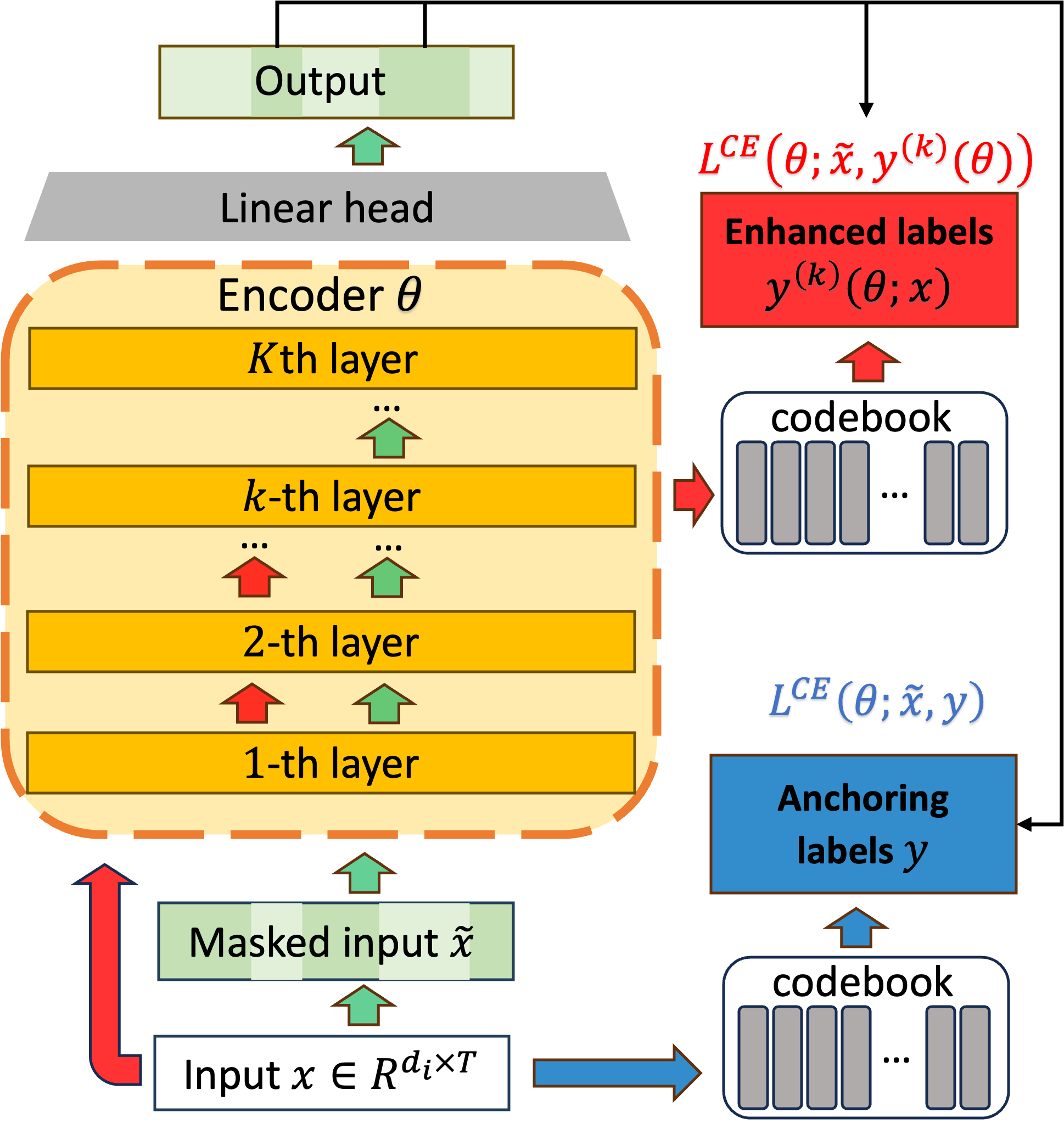}
\vspace{-0.2cm}
\caption{Overview of BiRQ: a bilevel SSL framework combining anchoring labels (lower-level) for stability and enhanced labels (upper-level) for improved representation learning. The encoder, parameterized by $\theta$, is assumed to be a generic multi-layer network (e.g., Transformer, Conformer). The \textcolor{SeaGreen}{green} arrows show the forward pass taking input $x$. The \textcolor{red}{red} arrows show the pipeline to generate enhanced labels $y^{(k)}(\theta;x)$, with intermediate representations taken from the first $k$ layers and quantized via a random-projection quantizer. The \textcolor{blue}{blue} arrows show a parallel process in which random quantization on the raw input $x$ generates anchoring labels $y(x)$.
}
\vspace{-0.4cm}
\label{fig:Self-Conditioned BEST-RQ}
\end{figure}

Existing SSL approaches for speech typically involve two key components: (1) generation of PLs and (2) representation learning with a masked prediction paradigm~\cite{devlin2019bert}.
For (1), the method used strongly affects performance, but often comes with efficiency tradeoffs.
For instance, wav2vec 2.0~\cite{baevski2020wav2vec} uses a multi-layer convolutional encoder and a product quantizer to discretize latent representations. HuBERT~\cite{hsu2021hubert} applies a separate acoustic unit discovery module (e.g., $k$-means clustering) to iteratively refine pseudo-labels from learned representations. 
More recent work in computer vision with joint embedding methods~\cite{assran2023self} similarly depends on complex label encoders. 
For (2), the choice of model architecture for representation learning is crucial for downstream task performance. Recent advances in ASR encoders, such as the Conformer~\cite{gulati2020Conformer}, have demonstrated strong recognition accuracy, yet they are not inherently equipped with modules for pseudo-label generation, making them less directly compatible with methods like wav2vec 2.0~\cite{baevski2020wav2vec} and HuBERT~\cite{hsu2021hubert}.

A more recent work, BEST-RQ~\cite{chiu2022self}, demonstrates strong performance on ASR benchmarks. It achieves efficiency in pseudo-label generation via a random quantizer and combines well with strong encoder architectures such as the Conformer~\cite{gulati2020Conformer}, which underpins many state-of-the-art ASR systems. However, insights from other SSL baselines show that the quality of pseudo-labels significantly impacts performance. For instance, HuBERT~\cite{hsu2021hubert} demonstrates that using a powerful label encoder and a multi-stage training process, which refines targets from the model's learned representations, can lead to substantial performance improvements.

Building on these insights, we improve upon BEST-RQ~\cite{chiu2022self} by integrating an iterative target refinement strategy~\cite{xu2020iterative,hsu2021hubert}, resulting in higher-quality pseudo-labels while preserving \textbf{end-to-end} learning as well as \textbf{computational} and \textbf{memory efficiency}. This leads to our proposed \textbf{Bi}level self-labeling \textbf{R}andom \textbf{Q}uantization (BiRQ) method for speech SSL, as illustrated in Figure~\ref{fig:Self-Conditioned BEST-RQ}.
Given a multi-layer acoustic encoder of general architecture and an unmasked speech input sample, BiRQ uses part of the encoder itself as a label encoder to generate an intermediate representation, which is then discretized with a random-projection quantizer to form the \emph{enhanced label}. The encoder is then trained, under the mask-and-predict paradigm, to reconstruct randomly masked regions of input using these labels.
To avoid degenerate solutions, an additional random quantizer generates an \emph{anchoring label} directly from the raw input. Training thus follows a Bi-Level Optimization (BLO) principle: the encoder learns from enhanced labels (upper-level) while staying consistent with anchoring labels (lower-level). See Section~\ref{sec: BiRO Method} for details.

Given trainable multi-layer encoder $\theta$ and input unlabeled data $x$, this BLO framework allows the encoder to leverage its own intermediate representations as pseudo labels $y^{(k)}(\theta;x)$ for supervision, while the raw-input labels $y(x)$, produced without any external clustering or auxiliary encoder, anchor training, and guard against collapse. 
To make the training procedure time-efficient and practical for large-scale problems, we develop a fully single-loop first-order optimization algorithm. As a by-product, since BiRQ reuses part of the model (e.g., layers $1$-$k$) as the label encoder, it does not need a separate pseudo-label encoder and is thus memory-efficient. Finally, the random-projection quantizer keeps label generation lightweight (no costly k-means or additional encoder) and, together with differentiable Gumbel-softmax selection, enables end-to-end training that propagates useful learning signals from the mask-and-predict objective while maintaining stability through the anchoring constraint. 

We evaluate BiRQ using various Conformer configurations and sizes on multiple datasets and find that BiRQ consistently outperforms BEST-RQ at similar computational cost and low complexity.

\section{Related Work}
\noindent\textbf{Self-supervised speech recognition.}
The earliest approaches to speech SSL used generative models trained with likelihood maximization~\cite{glarner2018full}.
Prediction-based self-supervised learning has recently gained more attention. Early methods exploited the sequential nature of speech, predicting future frames based on past observations~\cite{oord2018representation,schneider2019wav2vec}. 
Recently, random mask-and-predict training, which encourages the model to reconstruct masked portions of the input, has demonstrated strong performance and gained widespread interest~\cite{baevski2019effectiveness,baevski2020wav2vec,hsu2021hubert,chiu2022self}.
One line of work generates pseudo labels on learned continuous representations, requiring either a model architecture with a representation-learning module or external label encoders~\cite{baevski2019effectiveness,schneider2019wav2vec,baevski2020wav2vec,hsu2021hubert,fei2023jepa}.
However, such approaches often introduce additional computational and memory overhead, since training requires either maintaining a separate representation-learning module or an external label encoder, making them less compatible with recent ASR architectures such as the Conformer~\cite{gulati2020Conformer}.

\vspace{0.1cm}
\noindent\textbf{BLO for speech tasks.}
BLO has been extensively studied in the literature~\cite{bracken1973mathematical}. 
Recent progress in first-order BLO methods includes equilibrium propagation approaches~\cite{zucchet2022beyond} and penalty reformulations~\cite{shen2023penalty,kwon2023penalty,ye2023difference}. These first-order methods make BLO feasible for large-scale machine learning problems. Building on this, a growing body of single-loop algorithms further improves scalability and efficiency~\cite{jiang2025beyond,chen2024penalty}.
More recently, BLO has shown its effectiveness in speech-related applications, including unifying SSL with fine-tuning~\cite{zakarias2024bissl,saif2024joint} and sound event detection~\cite{zhang2024automated}. 

This paper builds on BEST-RQ~\cite{chiu2022self}, an efficient SSL method that generates labels via a fixed random-projection quantizer and a simple nearest-neighbor operation, without requiring a separate representation-learning module or external label encoder. BEST-RQ is flexible, compatible with acoustic encoders of various architectures, e.g., the Conformer~\cite{gulati2020Conformer}. While efficient, its performance can be further improved with more sophisticated labeling strategies, such as multiple codebooks~\cite{baevski2020wav2vec} or iterative refinement of learned representations~\cite{xu2020iterative,hsu2021hubert}, which motivates our work. In this paper, we apply BLO techniques to enhance BEST-RQ by incorporating iterative label refinement. Our proposed method, BiRQ, reuses the first $1$-$k$ layers of a multi-layer acoustic encoder as a label encoder, applying a random quantizer to generate \emph{enhanced pseudo labels} from the selected $k$-th intermediate representations for the upper-level BLO objective. The original BEST-RQ labels are retained as \emph{anchoring labels} for the lower-level BLO objective, ensuring stability during training.

\section{The BiRQ Method}
\label{sec: BiRO Method}
This section introduces BiRQ, which employs a self-labeling strategy by reusing part of the encoder to generate enhanced targets via a random quantizer, while anchoring targets from the raw input define a bilevel framework, enabling an end-to-end, computational and memory efficient SSL training pipeline. 

\subsection{Random-projector for pseudo label generations}
\label{sec: Random-projection Quantizer}
Let $x \in \mathbb{R}^{d_i \times T}$ denote the unmasked normalized input speech sequence from dataset $\mathcal{D}_{\text{tr}}$, where $d_i$ is the feature dimension and $T$ the number of time frames; and let $\theta$ denote the parameters of a $K$-layer encoder. Define $z^{(k)}(\theta;x)=\text{Encoder}_{\theta_{1\ldots k}}(x)\in \mathbb{R}^{d_h\times T}$ as the normalized intermediate representation at layer $k$, where $d_h$ is the hidden dimension. BiRQ introduces two sets of pseudo labels: an anchoring label $y(x)$ generated directly from $x$, and an enhanced label $y^{(k)}(\theta;x)$ generated from $z^{(k)}(\theta;x)$,
which are detailed next. 

\vspace{0.1cm}
\noindent\textbf{Generating anchoring label.} 
Let $C \in \mathbb{R}^{d_c \times N}$ denote an $N$-entry randomly generated fixed codebook with code entries of dimension $d_c$. Following BEST-RQ~\cite{chiu2022self}, given an input sample $x$, a fixed random projection matrix $P_{\text{anchor}} \in  \mathbb{R}^{d_i \times d_c}$ is adopted on $x$:
\begin{align}
    u(x) = P_{\text{anchor}}^\top x \in \mathbb{R}^{d_c \times T}
\end{align}
to match the dimension of the codebook.
In this way, for any time frame $t\in[T]$, the discrete anchoring label $[y(x)]_t$ can be generated by nearest-neighbor assignment:
\begin{align}
    [y(x)]_t = \text{OneHot}\Big(\arg\min_{n} \| [u(x)]_t - C_n \|^2, N\Big). \label{eq: onehot}
\end{align}
Here, $\text{OneHot}(k, N) \in \{0,1\}^N$ denotes the one-hot vector of length $N$ with a $1$ in the $k$-th position and $0$ elsewhere.

\vspace{0.1cm}
\noindent\textbf{Generating enhanced label.} BiRQ treats the first $k$ layers of the acoustic encoder as a label encoder to generate enhanced pseudo labels. Based on observations in pilot experiments, a rule of thumb is to choose $k \approx 0.7K$. For example, for a 5-layer encoder, choose $k=3$, and for a 10-layer encoder, choose $k=7$.  
BiRQ feeds $x$ to the encoder until the $k$-th layer to obtain normalized intermediate representation $z^{(k)}(\theta;x) \in \mathbb{R}^{d_h \times T}$. Then, it similarly adopts a fixed random projection matrix $P_{\text{enhance}} \in  \mathbb{R}^{d_h \times d_c}$ to get
\begin{align}
    u^{(k)}(\theta;x) = P_{\text{enhance}}^\top z^{(k)}(\theta;x) \in \mathbb{R}^{d_c \times T}. \label{eq: intermediate proj}
\end{align}
Notably, as $u^{(k)}(\theta;x)$ depends on the acoustic encoder model $\theta$, we want to generate labels that are differentiable to allow for future training. To achieve this, we apply Gumbel Softmax~\cite{jang2017categorical} as a differentiable surrogate to the nearest-neighbor principle:
\begin{align}
    [y^{(k)}(\theta;x)]_t 
    = \text{SoftMax}\!\left(-\frac{\|[u^{(k)}(\theta; x)]_t - C_n \|^2 + v_n}{\tau}\right),
    \label{eq: softmax}
\end{align}
where $v_n = -\ln(-\ln(q_n))$, $q_n \sim \text{Unif}(0,1)$, and $\tau > 0$ is a temperature parameter (set to $0.5$ in our experiments). 
The operator $\text{SoftMax}(\cdot)$ takes an $N$-dimensional score vector as input and outputs an $N$-dimensional categorical distribution, 
so that $[y^{(k)}(\theta;x)]_t$ serves as a differentiable relaxation of the discrete one-hot label.

\subsection{Bilevel SSL with anchoring and enhanced labels}
\label{sec: Learning lower-level from anchoring label}

BiRQ adopts a BERT-style~\cite{devlin2019bert} mask-and-predict scheme for SSL training using the two sets of generated labels in Sec.~\ref{sec: Random-projection Quantizer} to construct bilevel objectives.  
Given input $x$, a subset of input frames $M \subseteq [T]$ is randomly masked to obtain $\tilde{x}$, i.e., $x[M] = \text{GaussianNoise}(0,0.1)$ following BEST-RQ~\cite{chiu2022self}. The masked sequence $\tilde{x}$ is then passed through the encoder $\theta$, and a linear layer on top of the encoder produces output logits $o(\theta;\tilde{x}) \in \mathbb{R}^{N \times T}$ to match the codebook dimension.  

The cross-entropy loss for any set of pseudo-labels $y$ is defined on the masked positions $M$ as:
\begin{align}
\mathcal{L}_{\text{CE}}(\theta; \tilde{x}, y) = - \sum_{t \in M} [y]_t^\top \log \text{SoftMax}([o(\theta;\tilde{x})]_t)
\end{align}
which accommodates both the anchoring labels $y(x)$ (independent of $\theta$) and the enhanced labels $y^{(k)}(\theta;x)$ (dependent on $\theta$).

\vspace{0.1cm}
\noindent\textbf{Lower-level objective.} Using the anchoring labels, the lower-level (LL) objective is:
\begin{equation}
    G(\theta) := \frac{1}{|\mathcal{D}_{\text{tr}}|} \sum_{x \in \mathcal{D}_{\text{tr}}} 
    \mathcal{L}_{\text{CE}}(\theta; \tilde{x}, y(x)), \label{eq: G theta}
\end{equation}
which corresponds to the standard SSL objective used in BEST-RQ~\cite{chiu2022self}. The LL objective stabilizes training by providing a reliable pseudo-label supervision signal independent of $\theta$.

\vspace{0.1cm}
\noindent\textbf{Upper-level objective.}  
The upper-level (UL) objective leverages enhanced pseudo-labels generated from intermediate encoder representations, allowing the model to refine its predictions based on its evolving features:
\begin{equation}
    F(\theta) := \frac{1}{|\mathcal{D}_{\text{tr}}|} \sum_{x \in \mathcal{D}_{\text{tr}}} 
    \mathcal{L}_{\text{CE}}(\theta; \tilde{x}, y^{(k)}(\theta; x)). \label{eq: F theta}
\end{equation}
Because $y^{(k)}(\theta;x)$ is differentiable in $\theta$, the UL objective encourages end-to-end label refinement from representation learning.

\vspace{0.1cm}
\noindent\textbf{Bilevel formulation.}  
Combining the LL and UL objectives naturally leads to a simple BLO problem, where we require the lower-level objective to be learned near-optimally:
\begin{equation}
\min_{\theta \in S_\delta(\theta)}  \; F(\theta)~~~\text{s.t. }~~ S_\delta(\theta):= \big\{ \theta \;\big|\; G(\theta) - \min_{\theta'} G(\theta') \le \delta \big\}. \label{eq: BLO}
\end{equation}
Here, the LL problem defines a feasible set $S_\delta(\theta)$ for the UL problem, consisting of parameters that achieve near-optimal performance on the anchoring labels.

\subsection{Penalty-based single-loop algorithm}
\label{sec: BLO algorithm}

To solve the bilevel optimization problem in \eqref{eq: BLO}, we adopt a penalty-based single-level reformulation~\cite{shen2023penalty,kwon2023penalty}:
\begin{equation}
\min_\theta w_1 F(\theta) + w_2 \big(G(\theta) - \min_{\theta'} G(\theta') \big), \label{eq: penalty BLO}
\end{equation}
where $w_2 / w_1$ is called the penalty constant and is usually taken as a large constant, 
and $\min_{\theta'} G(\theta')$ is a constant and can therefore be omitted. This formulation encourages the model to simultaneously optimize the enhanced-label (UL) objective $F(\theta)$ while keeping the LL objective $G(\theta)$ close to optimal with the penalty constant $w_2 /w_1$ effectively controlling the LL relaxation $\delta$ in \eqref{eq: BLO}.  
The equivalence between the original bilevel problem \eqref{eq: BLO} and the penalty reformulation \eqref{eq: penalty BLO} holds under mild assumptions~\cite{chen2025foops,chen2024penalty}.  
\begin{lemma}[\cite{chen2024penalty}]
\label{lemma: equivalence of penalty BLO} 
If $\cup_{x \in S_0} \partial F(x)$ is bounded and $G(\theta) - \min_\theta G$ satisfies a Hölderian error bound with exponent $\alpha>1$, then setting $w_2/w_1=\Omega(\delta^{1-\alpha})$ ensures that any $\epsilon$-optimal solution of \eqref{eq: penalty BLO} is also an $\epsilon$-optimal solution of \eqref{eq: BLO}.
\end{lemma}
In our setting, the gradient of $F$ is bounded by the cross-entropy with Softmax outputs, and $G$ is either strongly convex ($\alpha=2$) or admits a local error bound induced by anchoring labels, satisfying Lemma~\ref{lemma: equivalence of penalty BLO}.
After reformulation, the bilevel problem reduces to a single differentiable objective in $\theta$, enabling standard first-order optimization in a single loop and avoiding the nested iterations typical of classical bilevel solvers~\cite{shen2023penalty,kwon2023penalty}.
This extends the standard SSL objective in BEST-RQ~\cite{chiu2022self} by incorporating differentiable enhanced labels, while the near-optimal LL objective provides stable supervision, yielding an efficient end-to-end SSL training pipeline. The implementation is given in Algorithm~\ref{alg:birq}.

\begin{algorithm}[t]
\label{alg: BLO}
\caption{BiRQ Training for Speech SSL}
\label{alg:birq}
\KwIn{Speech dataset $\mathcal{D}_{\text{tr}}$, acoustic encoder $\theta$, codebook $C$, intermediate layer $k$, bilevel parameter $w_1,w_2$, learning rate $\eta$ 
}
\For{each minibatch $x \sim \mathcal{D}_{\text{tr}}$}{
\tcp{Generate labels}
Generate anchor labels $y(x)$ by nearest-neighbor in \eqref{eq: onehot}\;
Extract normalized $k$-th layer latent features $z^{(k)}$ and project it to $u^{(k)}(\theta;x)$ following \eqref{eq: intermediate proj}\;
Construct enhanced labels $y^{(k)}(\theta;x)$ following \eqref{eq: softmax}\;

\tcp{Forward pass}
Compute LL loss $G(\theta; y(x))$ in \eqref{eq: G theta}\;
Compute UL loss $F(\theta; y^{(k)}(\theta;x))$ in \eqref{eq: F theta}\;

\tcp{Parameter update}
Update encoder: $\theta \leftarrow \theta - \eta (w_1\nabla F(\theta) + w_2\nabla G(\theta))$\;
}
\KwOut{Trained encoder parameters $\theta$}
\end{algorithm}

\section{Experiments}

\begin{table}[t]
    \centering
    \begin{tabular}{l|c|c}
        \hline \hline
        Method & test-clean & test-other  \\
        \hline \hline
        \multicolumn{3}{c}{\textbf{Main Comparison}} \\
        \hline
        Sup. Baseline (ep=100) & 8.4\% & 24.4\% \\
        BEST-RQ~\cite{chiu2022self} (ep=100) & 7.1\% & 20.5\% \\
        ~~~ + iter. relabeling (ep=100) & 6.3\% & 18.6\% \\
        ~~~ + iter. relabeling$^{2}$ (ep=100) & 6.2\% & 18.7\% \\
        \textbf{BiRQ (k=3, ep=100)} & 6.6\% & 19.1\% \\
        ~~~ \textbf{BiRQ (ep=200)} & \textbf{6.1\%} & \textbf{17.4\%} \\
        ~~~ \textbf{BiRQ (ep=300)} & \textbf{5.9\%} & \textbf{17.2\%} \\
        \hline \hline
        \multicolumn{3}{c}{\textbf{Ablation Study: Intermediate Layer Choice \(k\)}} \\
        \hline
        BiRQ (ep100, k=2) & 7.4\% & 21.1\% \\
        BiRQ (ep100, k=4) & 6.4\% & 17.9\% \\
        \hline \hline
        \multicolumn{3}{c}{\textbf{Variant / Extended Settings}} \\
        \hline
        BiRQ-4CB (ep100, k=3) &\textbf{6.2\% }& \textbf{16.3\%} \\
        \hline \hline
    \end{tabular}
    \vspace{-0.2cm}
    \caption{WERs of 5-layer 137M Conformer (C1), 
    fine-tuned on 100h Librispeech labeled logmel data. SSL pretraining of BEST-RQ and BiRQ is on 960h Librispeech.
    \textbf{Main comparison:} BiRQ versus the non-SSL supervised-only baseline, BEST-RQ~\cite{chiu2022self}, 
    and a HuBERT-style iterative labeling refinement variant. 
    \textbf{Ablation study:} Effect of different intermediate layers \(k\) 
    for generating enhanced pseudo-labels. 
    \textbf{Variant / extended Settings:} Use
    multi-codebook configurations. Best results are highlighted in bold.
    }
    \label{tab:birq_performance}
    \vspace{-0.3cm}
\end{table}


\noindent\textbf{Datasets.} 
We evaluate our method on multiple speech corpora. In one experiment we use the \textit{LibriSpeech} corpus~\cite{panayotov2015librispeech}, with 960 hours of unlabeled audio for SSL pretraining and the 100-hour labeled subset for fine-tuning and evaluation. Evaluation is reported on the official \texttt{test-clean} and \texttt{test-other} splits using word error rate (WER).  In another experiment we use a subset of the \textbf{YODAS} corpus~\cite{li2024yodas}, consisting of 5,000 hours of randomly selected unlabeled conversational audio for SSL. Fine-tuning is performed on the \textbf{AMI} meeting dataset~\cite{carletta2006announcing} using 150-hour labeled data, with WERs reported following \texttt{ami-ihm} / \texttt{ami-sdm} splits.  
All speech signals have a 16 kHz sampling rate. The acoustic features are 80-dimensional log-Mel filter banks computed using a 25 ms window with a 10 ms shift. 

\vspace{0.1cm}
\noindent\textbf{Self-supervised pretraining.} 
We pretrain BEST-RQ~\cite{chiu2022self} baselines and BiRQ using the \textbf{Conformer}~\cite{gulati2020Conformer} encoder. We experiment with three configurations: 
(\textbf{C1)} 5-layer, 1024-width, 8-head with attention window of 200 (137M parameters), 
\textbf{(C2)} 10-layer, 768-width, 6-head with full attention (155M parameters), and 
\textbf{(C3)} 10-layer, 1024-width, 8-head with attention window of 200 (275M parameters). For (C1), we also compare with a HuBERT-style multi-stage iterative labeling variant of BEST-RQ (denoted ``+iter. relabeling") where refined pseudo labels are generated offline using the representations from the intermediate layer ($3$rd-layer) from the previous BEST-RQ model for training of current model for 100 epochs.

Input acoustic features are log-mel spectrograms and are downsampled by stacking every two consecutive frames. Following~\cite{chiu2022self}, span masking is applied by randomly selecting $2\%$ of the total frame length and masking spans of 20 frames (effectively 40 frames due to stacking). For BiRQ, the loss weights are set to $w_1 = 0.1$ and $w_2 = 2.4$, corresponding to a bilevel penalty parameter $\gamma = w_2/w_1=24$.  

The 137M and 155M models are trained for 100 epochs with batch sizes of 100 and 64, 
respectively, using the 960-hour LibriSpeech~\cite{panayotov2015librispeech} unlabeled audio dataset. Both models use the AdamW optimizer 
with learning rates of $2\times10^{-4}$ and $1\times10^{-4}$, respectively.  
The 275M model is trained for 100 epochs with batch sizes of 128, 
on the 5,000-hour YODAS speech using a learning rate of $1\times10^{-4}$ under otherwise identical settings.

\vspace{0.1cm}
\noindent\textbf{Supervised fine-tuning.} 
The 137M and 155M models are fine-tuned on the 100-hour labeled subset of LibriSpeech, while the 275M model is fine-tuned on the 150-hour labeled speech of the AMI meeting dataset. Fine-tuning is performed using the CTC loss~\cite{graves2006connectionist}. The Adam optimizer is employed with a 10-epoch linear warmup to a peak learning rate of 0.001, held for 10 epochs, followed by 10-epoch annealing with per-epoch decay of rate $\sqrt{2}$. Decoding of Librispeech is conducted using the $4$-gram external language model (LM) provided by the Librispeech dataset.  Decoding of AMI meetings is conducted using an in-house interpolated 4-gram external LM.

\vspace{0.1cm}
\noindent\textbf{137M Model Results.}
Table~\ref{tab:birq_performance} shows the performance of BiRQ compared to supervised baselines, BEST-RQ, and a HuBERT-style multi-stage iterative relabeling variant of BEST-RQ.
Our results demonstrate that BiRQ consistently surpasses the supervised-only baseline, BEST-RQ and its iterative relabeling variant, achieving $5.9\%$ WER on \texttt{test-clean} and $17.2\%$ on \texttt{test-other} after 300 epochs. This indicates that BiRQ attains the benefits of multiple iterative refinements within a single, unified bilevel formulation, while using only half the memory footprint compared to iterative relabeling.
Ablation studies show that $k=3$ yields the strongest results, supporting our rule of thumb that $k \approx 0.7K$.
In addition, using multiple codebooks (4 vs. 1) further improves performance, confirming the importance of pseudo-label quality and diversity. 

\vspace{0.1cm}
\vspace{0.1cm}
\noindent\textbf{Scaling to 155M and 275M.}
Table~\ref{tab:librispeech_results} reports results on a thinner but deeper 155M Conformer. BiRQ continues to deliver strong improvements, reducing WERs to $5.0\% / 12.6\%$ compared to $6.8\% / 19.6\%$ for BEST-RQ. This confirms that BiRQ generalizes well to different architectural configurations. Moreover, this setup incorporates full attention, showing that simple but more powerful architectural adjustments, similar to increasing codebooks, can complement BiRQ.
In Table~\ref{tab:ami_results}, scaling further to a 275M Conformer pretrained on the 5k-hour YODAS dataset shows that BiRQ provides consistent gains, lowering \texttt{ami-ihm/-sdm} WERs to $16.3\% / 34.0\%$ from $18.4\% / 37.2\%$ with BEST-RQ.
These results confirm that BiRQ not only improves label quality at modest model sizes but also scales effectively to larger architectures and datasets, making it a versatile framework for self-supervised speech representation learning.

\begin{table}[t]
    \centering
    \begin{tabular}{l|c|c}
        \hline \hline
        Method & test-clean & test-other \\
        \hline \hline
        Sup. Baseline (ep=100) & 7.4\% & 20.5\% \\
        BEST-RQ~\cite{chiu2022self} (ep=100) & 6.8\% & 19.6\% \\
        \textbf{BiRQ (ep=100, k=7)} & \textbf{5.0\%} & \textbf{12.6\%} \\
        \hline \hline
    \end{tabular}
    \vspace{-0.2cm}
    \caption{WERs of 10-layer 155M Conformer (C2), fine-tuned on 100h Librispeech labeled logmel data. SSL pretrainings of BEST-RQ and BiRQ are on 960h Librispeech.}
    \label{tab:librispeech_results}
\end{table}

\begin{table}[t]
    \centering
    \begin{tabular}{l|c|c}
        \hline \hline
        Method & ami-ihm & ami-sdm \\
        \hline \hline
        Sup. Baseline (ep=100) & 25.7\% & 47.0\% \\
        BEST-RQ~\cite{chiu2022self} (ep=100) & 18.4\% & 37.2\% \\
        \textbf{BiRQ (ep=100, k=7)} & \textbf{16.3\%} & \textbf{34.0\%} \\
        \hline \hline
    \end{tabular}
    \vspace{-0.1cm}
    \caption{WERs of 10-layer 275M Conformer (C3), fine-tuned on 150h AMI Meeting labeled logmel data. SSL pretrainings of BEST-RQ and BiRQ are on 5k Yodas.}
    \label{tab:ami_results}
    \vspace{-0.3cm}
\end{table}

\section{Concluding Remarks}

This paper introduces a simple yet effective bilevel framework - BiRQ for self-supervised speech representation learning, leveraging both stable anchoring labels and differentiable enhanced labels derived from intermediate encoder outputs. Our experiments show that this approach consistently improves performance over existing methods, including standard BEST-RQ and iterative offline label updates, across multiple model scales and datasets. Ablation studies further highlight the importance of the intermediate layer choice and the potential of multi-codebook configurations.

\bibliographystyle{IEEEbib}
\bibliography{reference,bilevel_ref}

\begin{thebibliography}{10}

\bibitem{hsu2021hubert}
Wei-Ning Hsu, Benjamin Bolte, Yao-Hung~Hubert Tsai, Kushal Lakhotia, Ruslan Salakhutdinov, and Abdelrahman Mohamed,
\newblock ``Hubert: Self-supervised speech representation learning by masked prediction of hidden units,''
\newblock {\em IEEE/ACM transactions on audio, speech, and language processing}, vol. 29, pp. 3451--3460, 2021.

\bibitem{chiu2022self}
Chung-Cheng Chiu, James Qin, Yu~Zhang, Jiahui Yu, and Yonghui Wu,
\newblock ``Self-supervised learning with random-projection quantizer for speech recognition,''
\newblock in {\em International Conference on Machine Learning}, 2022, pp. 3915--3924.

\bibitem{devlin2019bert}
Jacob Devlin, Ming-Wei Chang, Kenton Lee, and Kristina Toutanova,
\newblock ``Bert: Pre-training of deep bidirectional transformers for language understanding,''
\newblock in {\em Conference of the North American chapter of the association for computational linguistics: human language technologies}, 2019, pp. 4171--4186.

\bibitem{baevski2020wav2vec}
Alexei Baevski, Yuhao Zhou, Abdelrahman Mohamed, and Michael Auli,
\newblock ``wav2vec 2.0: A framework for self-supervised learning of speech representations,''
\newblock {\em Advances in neural information processing systems}, vol. 33, pp. 12449--12460, 2020.

\bibitem{li2023blip}
Junnan Li, Dongxu Li, Silvio Savarese, and Steven Hoi,
\newblock ``Blip-2: Bootstrapping language-image pre-training with frozen image encoders and large language models,''
\newblock in {\em International conference on machine learning}. PMLR, 2023, pp. 19730--19742.

\bibitem{assran2023self}
Mahmoud Assran, Quentin Duval, Ishan Misra, Piotr Bojanowski, Pascal Vincent, Michael Rabbat, Yann LeCun, and Nicolas Ballas,
\newblock ``Self-supervised learning from images with a joint-embedding predictive architecture,''
\newblock in {\em Proceedings of the IEEE/CVF Conference on Computer Vision and Pattern Recognition}, 2023, pp. 15619--15629.

\bibitem{gulati2020Conformer}
Anmol Gulati, James Qin, Chung-Cheng Chiu, Niki Parmar, Yu~Zhang, Jiahui Yu, Wei Han, Shibo Wang, Zhengdong Zhang, Yonghui Wu, et~al.,
\newblock ``Conformer: Convolution-augmented transformer for speech recognition,''
\newblock {\em arXiv preprint arXiv:2005.08100}, 2020.

\bibitem{xu2020iterative}
Qiantong Xu, Tatiana Likhomanenko, Jacob Kahn, Awni Hannun, Gabriel Synnaeve, and Ronan Collobert,
\newblock ``Iterative pseudo-labeling for speech recognition,''
\newblock {\em arXiv preprint arXiv:2005.09267}, 2020.

\bibitem{glarner2018full}
Thomas Glarner, Patrick Hanebrink, Janek Ebbers, and Reinhold Haeb-Umbach,
\newblock ``Full bayesian hidden markov model variational autoencoder for acoustic unit discovery.,''
\newblock in {\em INTERSPEECH}, 2018.

\bibitem{oord2018representation}
Aaron van~den Oord, Yazhe Li, and Oriol Vinyals,
\newblock ``Representation learning with contrastive predictive coding,''
\newblock {\em arXiv preprint arXiv:1807.03748}, 2018.

\bibitem{schneider2019wav2vec}
Steffen Schneider, Alexei Baevski, Ronan Collobert, and Michael Auli,
\newblock ``wav2vec: Unsupervised pre-training for speech recognition,''
\newblock {\em arXiv preprint arXiv:1904.05862}, 2019.

\bibitem{baevski2019effectiveness}
Alexei Baevski, Michael Auli, and Abdelrahman Mohamed,
\newblock ``Effectiveness of self-supervised pre-training for speech recognition,''
\newblock {\em arXiv preprint arXiv:1911.03912}, 2019.

\bibitem{fei2023jepa}
Zhengcong Fei, Mingyuan Fan, and Junshi Huang,
\newblock ``A-jepa: Joint-embedding predictive architecture can listen,''
\newblock {\em arXiv preprint arXiv:2311.15830}, 2023.

\bibitem{bracken1973mathematical}
Jerome Bracken and James~T McGill,
\newblock ``Mathematical programs with optimization problems in the constraints,''
\newblock {\em Operations Research}, vol. 21, no. 1, pp. 37--44, 1973.

\bibitem{zucchet2022beyond}
Nicolas Zucchet and Jo{\~a}o Sacramento,
\newblock ``Beyond backpropagation: bilevel optimization through implicit differentiation and equilibrium propagation,''
\newblock {\em Neural Computation}, vol. 34, no. 12, pp. 2309--2346, 2022.

\bibitem{shen2023penalty}
Han Shen, Quan Xiao, and Tianyi Chen,
\newblock ``On penalty-based bilevel gradient descent method,''
\newblock in {\em Proc. International Conference on Machine Learning}, Honolulu, HI, 2023.

\bibitem{kwon2023penalty}
Jeongyeol Kwon, Dohyun Kwon, Steve Wright, and Robert Nowak,
\newblock ``On penalty methods for nonconvex bilevel optimization and first-order stochastic approximation,''
\newblock in {\em Proc. International Conference on Learning Representations}, 2024.

\bibitem{ye2023difference}
Jane~J Ye, Xiaoming Yuan, Shangzhi Zeng, and Jin Zhang,
\newblock ``Difference of convex algorithms for bilevel programs with applications in hyperparameter selection,''
\newblock {\em Mathematical Programming}, vol. 198, no. 2, pp. 1583--1616, 2023.

\bibitem{jiang2025beyond}
Liuyuan Jiang, Quan Xiao, Lisha Chen, and Tianyi Chen,
\newblock ``Beyond value functions: Single-loop bilevel optimization under flatness conditions,''
\newblock {\em arXiv preprint arXiv:2507.20400}, 2025.

\bibitem{chen2024penalty}
Pengyu Chen, Xu~Shi, Rujun Jiang, and Jiulin Wang,
\newblock ``Penalty-based methods for simple bilevel optimization under h$\backslash$"$\{$o$\}$ lderian error bounds,''
\newblock {\em arXiv preprint arXiv:2402.02155}, 2024.

\bibitem{zakarias2024bissl}
Gustav~Wagner Zakarias, Lars~Kai Hansen, and Zheng-Hua Tan,
\newblock ``Bissl: Enhancing the alignment between self-supervised pretraining and downstream fine-tuning via bilevel optimization,''
\newblock {\em arXiv preprint arXiv:2410.02387}, 2024.

\bibitem{saif2024joint}
AFM Saif, Xiaodong Cui, Han Shen, Songtao Lu, Brian Kingsbury, and Tianyi Chen,
\newblock ``Joint unsupervised and supervised training for automatic speech recognition via bilevel optimization,''
\newblock {\em arXiv preprint arXiv:2401.06980}, 2024.

\bibitem{zhang2024automated}
Wenjie Zhang, Peng Yu, Jun Yin, Xiaoheng Jiang, and Mingliang Xu,
\newblock ``Automated audio data augmentation network using bi-level optimization for sound event localization and detection,''
\newblock {\em IEEE Signal Processing Letters}, 2024.

\bibitem{jang2017categorical}
Eric Jang, Shixiang Gu, and Ben Poole,
\newblock ``Categorical reparameterization with gumbel-softmax,''
\newblock in {\em International Conference on Learning Representations}, 2017.

\bibitem{chen2025foops}
Lisha Chen, Quan Xiao, Ellen~Hidemi Fukuda, Xinyi Chen, Kun Yuan, and Tianyi Chen,
\newblock ``Efficient first-order optimization on the pareto set for multi-objective learning under preference guidance,''
\newblock in {\em International Conference on Machine Learning}, 3 2025.

\bibitem{panayotov2015librispeech}
Vassil Panayotov, Guoguo Chen, Daniel Povey, and Sanjeev Khudanpur,
\newblock ``Librispeech: an asr corpus based on public domain audio books,''
\newblock in {\em IEEE International Conference on Acoustics, Speech, and Signal Processing}, 2015.

\bibitem{li2024yodas}
Xinjian Li, Shinnosuke Takamichi, Takaaki Saeki, William Chen, Sayaka Shiota, and Shinji Watanabe,
\newblock ``Yodas: Youtube-oriented dataset for audio and speech,''
\newblock in {\em 2023 IEEE Automatic Speech Recognition and Understanding Workshop (ASRU)}. IEEE, 2023, pp. 1--8.

\bibitem{carletta2006announcing}
Jean Carletta,
\newblock ``Announcing the ami meeting corpus,''
\newblock {\em The ELRA Newsletter}, vol. 11, no. 1, pp. 3--5, 2006.

\bibitem{graves2006connectionist}
Alex Graves, Santiago Fern{\'a}ndez, Faustino Gomez, and J{\"u}rgen Schmidhuber,
\newblock ``Connectionist temporal classification: labelling unsegmented sequence data with recurrent neural networks,''
\newblock in {\em International Conference on Machine Learning}, 2006.

\end{thebibliography}

\end{document}